\documentclass[letterpaper]{article} 
\usepackage{aaai2026}
\usepackage{times}  
\usepackage{helvet}  
\usepackage{courier}  
\usepackage[hyphens]{url}  
\usepackage{graphicx} 
\usepackage{natbib}  
\usepackage{caption} 
\usepackage{algorithm}
\usepackage{algorithmic}

\usepackage{xcolor}
\usepackage{booktabs}

\usepackage{multirow}
\usepackage{amssymb}
\usepackage{gradient-text}
\usepackage{subcaption}
\usepackage{pifont}
\usepackage[breakable]{tcolorbox}
\usepackage{enumitem}
\usepackage{colortbl}

\usepackage{listings}
\newcommand\cmark{\ding{52}}
\newcommand\xmark{\ding{55}}
\newcommand{\corresauth}{\textsuperscript{\dag}}

\usepackage{newfloat}
\usepackage{listings}
\DeclareCaptionStyle{ruled}{labelfont=normalfont,labelsep=colon,strut=off} 
\floatstyle{ruled}
\newfloat{listing}{tb}{lst}{}
\floatname{listing}{Listing}
\definecolor{promptBlueBack}{HTML}{EBF5FF}
\definecolor{promptBlueFrame}{HTML}{003366}
\definecolor{promptGreenBack}{HTML}{F0FFF0}
\definecolor{promptGreenFrame}{HTML}{004D00}
\definecolor{promptOrangeBack}{HTML}{FFF5E6}
\definecolor{promptOrangeFrame}{HTML}{D35400} 
\definecolor{mylightgray}{gray}{0.9}

\newtcolorbox{systempromptbox}[1]{
    colback=promptBlueBack,   
    colframe=promptBlueFrame, 
    title=#1,                 
    fonttitle=\bfseries, 
    boxrule=0.5mm, 
    arc=2mm, 
    left=2mm,
    right=2mm, 
    top=2mm,
    bottom=2mm,
    breakable,
    width=\textwidth,
}

\newtcolorbox{userpromptbox}[1]{
    colback=promptGreenBack,    
    colframe=promptGreenFrame,  
    title=#1,             
    fonttitle=\bfseries, 
    boxrule=0.5mm, 
    arc=2mm, 
    left=2mm,
    right=2mm, 
    top=2mm,
    bottom=2mm,
    breakable,
}

\newtcolorbox{examplebox}[1]{
    colback=promptOrangeBack,    
    colframe=promptOrangeFrame,  
    title=#1,             
    fonttitle=\bfseries, 
    boxrule=0.5mm, 
    arc=2mm, 
    left=2mm,
    right=2mm, 
    top=2mm,
    bottom=2mm,
    breakable,
}

\title{How Foundational Skills Influence VLM-based Embodied Agents: \\ A Native Perspective}

\author{
    Bo Peng\textsuperscript{\rm 1,2}\equalcontrib, Pi Bu\textsuperscript{\rm 2}\equalcontrib, Keyu Pan\textsuperscript{\rm 4}, Xinrun Xu\textsuperscript{\rm 2,3}, Yinxiu Zhao\textsuperscript{\rm 2}, \\
    Miao Chen\textsuperscript{\rm 2}, Yang Du\textsuperscript{\rm 2}, Lin Li\textsuperscript{\rm 2}, Jun Song\textsuperscript{\rm 2}\corresauth, Tong Xu\textsuperscript{\rm 1}\thanks{Email: jsong.sj@alibaba-inc.com, tongxu@ustc.edu.cn}
}
\affiliations{
    \textsuperscript{\rm 1}University of Science and Technology of China\\
    \textsuperscript{\rm 2}Alibaba Group\\
    \textsuperscript{\rm 3}Institute of Software, Chinese Academy of Sciences \\
    \textsuperscript{\rm 4}Independent Researcher \\
}

\usepackage{bibentry}

\begin{document}

\maketitle

\begin{abstract}
Recent advances in vision–language models (VLMs) have shed light on human-level embodied intelligence. However, existing benchmarks for VLM-driven embodied agents still rely on high-level commands or discretised action spaces—``non-native'' settings that diverge markedly from the real world. Moreover, current benchmarks focus exclusively on high-level tasks,  while lacking joint evaluation and analysis on both low- and high-level. To bridge these gaps, we present \textbf{NativeEmbodied}, a challenging benchmark for VLM-driven embodied agents that adopts a unified, native low-level action space. Built upon diverse simulated scenes, NativeEmbodied first designs three representative high-level tasks in complex scenarios to evaluate overall performance. For more detailed and comprehensive performance analysis, we further decouple the entangled skills behind complex tasks and construct four types of low-level tasks, each corresponding to a key fundamental embodied skill.
This joint evaluation across task and skill granularities enables a fine-grained assessment of embodied agent. Comprehensive experiments on the best VLMs reveal pronounced deficiencies in certain fundamental embodied skills. Further analysis shows that these bottlenecks severely constrain performance on high-level tasks. Our NativeEmbodied not only pinpoints the key challenges faced by current VLM-driven embodied agents, but also provides valuable insight for future development of this field. 
\end{abstract}

\begin{links}
\link{Code \& Datasets}{https://github.com/LivingFutureLab/NativeEmbodied}
\end{links}

\section{Introduction}

Recent advances in Vision-Language Models (VLMs) have catalyzed significant progress in embodied intelligence \cite{wang2024qwen2vl}, bringing us closer to intelligent agents that can operate in the simulator or physical world \cite{cheang2025gr3,wang2025escapecraft,embodimentcollaboration2025openx,brohan2023rt2}. These VLM-based embodied agents, capable of perceiving the environment through visual inputs, and perform complex task following natural language instructions~\cite{chen2025combatvla,tan2024cradle,cao20254dspatialsurvey,long2025surveyei,junpeng2025mllm}.

However, a fundamental challenge persists: 
How can we assess whether these models truly possess the capability to function in the real world, and which fundamental skills bottleneck their performance? This question becomes particularly important as current evaluation benchmarks for embodied agent exhibit several limitations: 1) \textbf{Non-Native Action Space}: Recent benchmarks \cite{cheng2025embodiedeval,yang2025embodiedbench} attempt to deploy VLM-based agents in embodied simulators and evaluate them through interactive tasks. They typically abstract low-level actions into high-level commands or functions that the agent can invoke directly (e.g., “look at the apple”, “teleport to the desk”) - what we term the ``non-native'' setting. This abstraction emphasizes task reasoning and planning, while eclipsing critical embodied skills such as spatial alignment and navigation, leading to a considerable gap from real world. 2)  \textbf{Coupled Task Design}: Existing benchmarks focus on high-level tasks that entangle multiple foundational skills and measure model performance primarily by overall success rate. Such coarse-grained task formulation and evaluation hinder the diagnosis of skill-level bottlenecks, yielding assessments that are neither comprehensive nor sufficiently fine-grained.
Those limitation highlights two critical questions:

\begin{itemize}[leftmargin=*]
  \item Q1: Which foundational skills are truly essential for VLM-based embodied agents?
  \item Q2: How do these foundational skills affect the execution of higher-level tasks?
\end{itemize}

\begin{table*}[htbp]
\small
\centering
\begin{tabular}{l|rccccccccc}
\toprule
\textbf{BenchMark}  & \textbf{Size} & \textbf{Task Level} & \textbf{Fine-Grained} & \textbf{Multimodal}  & \textbf{Native} & \textbf{Decoupled}  \\ \midrule

ALFRED \cite{shridhar2020alfred}     & 3,062 & High  & \xmark   & \cmark & \xmark & \xmark \\

ALFWorld \cite{shridhar2020alfworld}    & 274 & High  & \xmark  & \xmark & \xmark & \xmark \\

VLMbench \cite{zheng2022vlmbench}     & 4,760 & Low  & \xmark   & \cmark & \cmark & \xmark \\

Behavior-1k \cite{li2023behavior}     & 1,000 & High  & \xmark   & \cmark & \xmark & \xmark \\

Lota-bench \cite{choi2024lota}    & 308 & High  & \xmark   & \xmark & \cmark & \xmark \\

GOAT-bench \cite{khanna2024goat}     & 3,919 & Low  & \xmark & \cmark & \cmark & \xmark \\

Embodied Agent Inferface \cite{li2024embodied}     & 438 & High  & \cmark   & \xmark & \xmark & \xmark \\

EmbodiedBench \cite{li2024embodied}     & 1,128 & High\&Low  & \cmark   & \cmark  & \xmark & \xmark \\

EmbodiedEval \cite{cheng2025embodiedeval}     & 328 & High & \xmark  & \cmark & \xmark & \xmark \\

\midrule
\textbf{NativeEmbodied (Ours)}     & 1,085 & High\&Low  & \cmark  & \cmark & \cmark & \cmark \\
\bottomrule
\end{tabular}
\caption{Comparisons between our NativeEmbodied and previous benchmarks . }
\label{tab:compare}
\end{table*}

To answer the above questions, we present \textbf{NativeEmbodied}, the first comprehensive benchmark that assesses VLMs’ multidimensional embodied skills from a native perspective. The following key features set NativeEmbodied apart from the other benchmarks: \textbf{1) Native Rollout Setting.} Built on AI2THOR~\cite{kolve2022ai2thor}—a widely used embodied simulator with richly detailed environments—NativeEmbodied adopts a native rollout setting. During a rollout, the agent receives only the initial task instruction, action history, and the egocentric images streamed by the simulator. In each turn, the agent are allowed to specify action only from AI2THOR's primitive action set, includes parameterizable rotations and movements. In this way, the agent is free to explore and interact with the environment in a native manner, making the benchmark more closely aligned with real-world conditions compared to previous ones.  \textbf{2) Decoupled Task Hierarchy.}  NativeEmbodied not only designs three categories of representative high-level tasks, but also decouples four categories of low-level tasks based on them. Each of these low-level tasks corresponds to a fundamental embodied skill. The synergistic evaluation from complex high-level tasks to decoupled low-level tasks facilitates more comprehensive and granular skill assessment and bottleneck analysis. 
Thereafter, we conducted extensive experiments and analyses with NativeEmbodied on 15 open-source and proprietary VLMs to explore the capabilities of existing embodied agents from a native perspective. 

Our contributions are summarized as follows:
\begin{itemize}[leftmargin=*]
\item We introduce a multidimensional, multigranular benchmark built upon native action spaces, providing a more realistic perspective for VLM-based embodied agents. 
\item We present a comprehensive evaluation system for fundamental embodied skills at a more raw and native level, where high- and low-level tasks are collaboratively evaluated to reveal skill-level bottlenecks, significantly enhancing the explainability of capability assessment. 
\item We provide extensive experimental validation across 15 open-source and closed-source models, offering valuable insights, with all resources and implementations publicly available to facilitate further research in this field.
\end{itemize}

\section{Related Work}
\subsubsection{Embodied Agent Benchmarks}
As shown in Table \ref{tab:compare}, recent years have witnessed a surge of benchmarks targeting vision-driven embodied agents, yet most remain domain-specific or modality-restricted. Classic benchmarks such as ALFWorld \cite{shridhar2020alfworld} and ALFRED \cite{shridhar2020alfred} focus on high-level household tasks but ignore low-level control; conversely, VLMbench \cite{zheng2022vlmbench} and GOAT-bench \cite{khanna2024goat} evaluate low-level manipulation and navigation, respectively, but are confined to isolated embodied skills. Concurrently, EmbodiedBench \cite{yang2025embodiedbench} introduces a multi-domain suite spanning household, manipulation, and navigation, while relying on high-level action when dealing with high-level tasks.
EmbodiedEval \cite{cheng2025embodiedeval} proposes a multi-domain benchmark for VLMs, yet its limited scale (328 instances) and absence of low-level tasks highlight the need for more comprehensive benchmarks.

\subsubsection{VLM-based Agents}
VLM-based agents typically ingest an interleaved sequence of images, text instructions, and optionally past actions, then output either free-form text or discrete/continuous action functions (i.e., non-native setting) that a downstream executor maps to low-level controls \cite{bai2023qwen,ui-tars,bai2025qwen25vl}. This paradigm has powered game agents \cite{xu2024survey} that generate controller commands from screen pixels and dialogue in Minecraft \cite{jucys2024vpt} and Pokémon \cite{hu2024pokellmon}, as well as Mobile agents that navigate mobiles to book flights \cite{lin2024showui,li2025hedgeagents,gu2025mobile}. When instantiated for embodied tasks, however, the agent must confront a native action space—open, close, pick up, and put down. In this paper, we hope the embodied agent can free to explore and interact with the environment in a native manner, making our NativeEmbodied benchmark more closely aligned with real-world conditions compared to previous ones.

\begin{figure*}[t]
  \centering
  \includegraphics[width=0.98\textwidth]{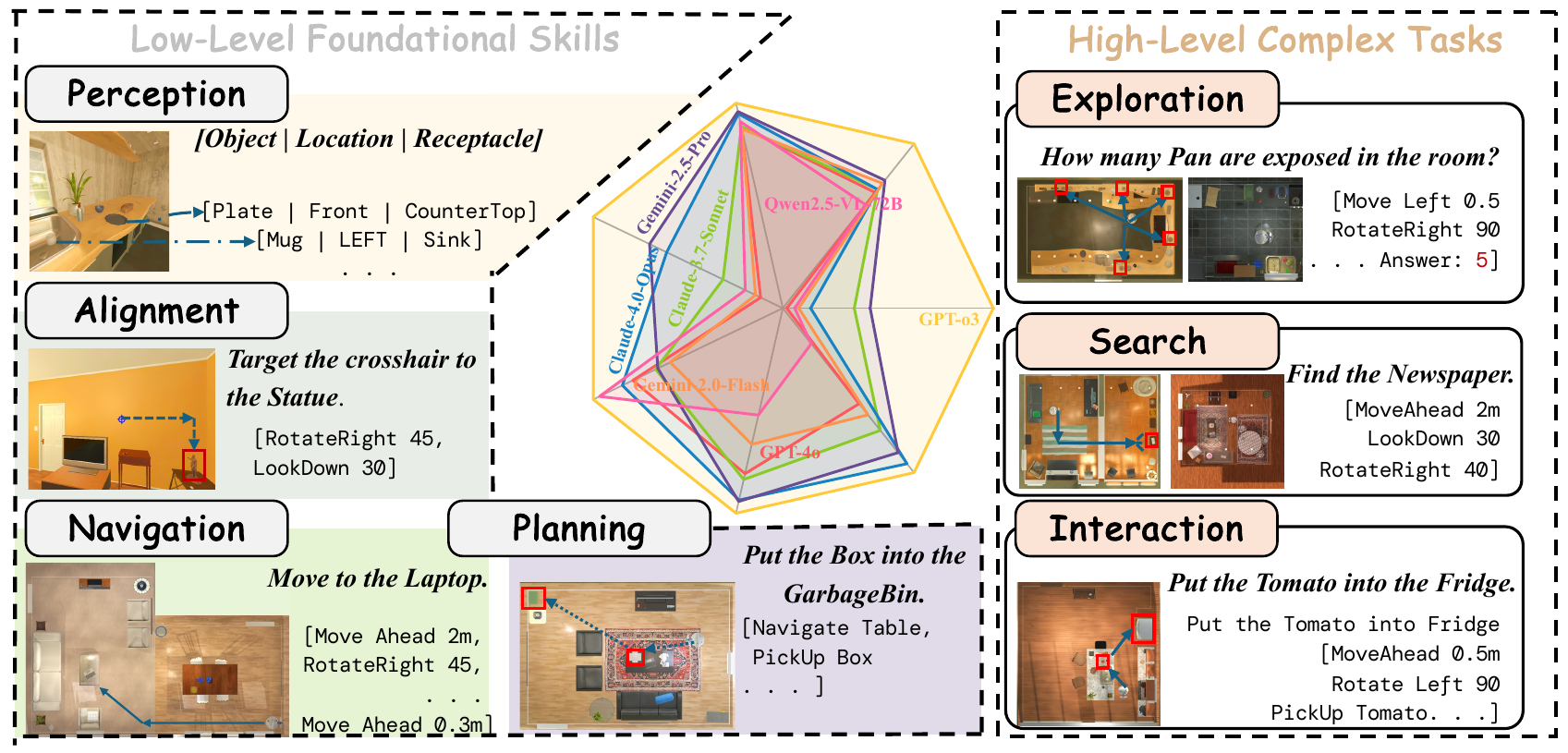}
  \caption{Our NativeEmbodied benchmark includes four fundamental low-level tasks (i.e., Perception, Alignment, Navigation and Planning) and three integrated high-level tasks (i.e., Exploration, Interaction, and Search).}
  \label{fig:single}
\end{figure*}

\section{NativeEmbodied Benchmark}
From a native perspective, we start with the native actions an agent can take. Specifically, we collect these basic moves and build a benchmark, NativeEmbodied, that checks four low-level tasks (e.g., center alignment and navigation). Because each subtask is separate and mix-and-match, we then combine high-level tasks (e.g., search). 
Through this bottom-up, decoupled setup, we enable analysis of the relationships between foundational capabilities and finial task success rates, revealing critical pathways of VLM-based embodied agent.

\subsection{Native Action Space}
To support the native setting, we define the native action space as follows:
\begin{itemize}[leftmargin=*]
\item MoveAhead x (meters): Move forward x meters
\item MoveBack x (meters): Move backward x meters
\item MoveLeft x (meters): Move left x meters
\item MoveRight x (meters): Move right x meters
\item RotateRight x (degrees): Rotate view right by x degrees
\item RotateLeft x (degrees): Rotate view left by x degrees
\item LookUp x (degrees): Tilt view upward by x degrees
\item LookDown x (degrees): Tilt view downward by x degrees
\end{itemize}

The nativeness of the action space lies in: unlike prior works using high-level functions like navigating\cite{cheng2025embodiedeval} or hardcoded action parameters\cite{yang2025embodiedbench}, it applies lowest-level action primitives (e.g. Move and Rotate) that a VLM can understand for multi-level tasks, combined with fully customized action parameters (e.g., distance, angle). These primitives and parameters capture the basic semantics of the agent-environment interaction, allowing agents to operate in the environment in a primitive and unconstrained manner. Our NativeEmbodied represents the first benchmark to provide completely unrestricted native action space across high-level tasks and low-level skills.

\subsection{High-level Complex Tasks}
We start with three representative high-level tasks that benchmark the overall performance in the native settings:

\noindent\textbf{Exploration.} This task poses questions related to objects in the environment, requiring agents to fully explore the environment to provide correct answers. For this task, we subdivide it into four subtypes:
\begin{itemize}
\item  \textit{Counting}: How many specified objects are exposed in the environment?
\item  \textit{Localization}: Which receptacle does the specified object locate in?
\item \textit{Receptacle Content}: Which of the following objects do (or don't) appear on the specified receptacle?
\item \textit{Co-existence}: Which of the following objects is (or isn't) on the same receptacle as the the specified object?
\end{itemize}

\noindent \textbf{Search.} This task requires agents to precisely locate and target specified objects within the environment. We overlay a crosshair at the center of the agent's egocentric obseravtion image to indicate the focal point.  The agent must approach the target object and align the crosshair with it to complete the task. This challenge demands that the agent not only identify the object's location but also navigate to it and execute fine-grained spatial alignment.

\noindent \textbf{Interaction.}
The task requires the agent to interact with objects in the scene to fulfill user instructions. Concretely, we focus on the representative pick-and-place scenario: the agent must place a specified object into a specified receptacle. The target object may be exposed in the environment or stored inside a closed receptacle, and the destination receptacle may be one that does not need to be opened (e.g., a tabletop) or one that does (e.g., a refrigerator). For this task we augment the original action space with four additional interaction primitives: PickUp, PutIn, Open, and Close.

\subsection{Low-level Foundational Skills}
While high-level tasks reveal an agent’s overall competence, they are not ideal for diagnosing specific skill deficiencies. The limitation is even starker in the native setting: here, a model’s core embodied abilities are tested directly, yet the multi-skill nature of the high-level tasks masks individual bottlenecks. For better evaluation, we therefore decompose the high-level tasks from a skill-centric perspective and introduce four classes of low-level tasks that each target a fundamental skills:

\noindent \textbf{Perception.} 
This task is designed to probe the agent’s perceptual abilities. The agent must describe the key semantic and spatial elements of the egocentric observation image in a predefined \textit{object}|\textit{location}|\textit{receptacle} triplet format:
\begin{itemize}[leftmargin=*]
\item List every object visible in the field of view.
\item Specify each object's spatial relationship to the agent.
\item Specify each object's receptacle.
\end{itemize}
This approach combines visual and spatial perception, and its structured format facilitates fine-grained evaluation of each aspect, making the evaluation in this paper more intuitive and flexible.

\noindent \textbf{Spatial Alignment.} Similar to the search task, the agent must align the center of its view with the specified object. However, to decouple from other foundational skills such as planning and navigation, we initialize the agent close to the target so the object is already within its egocentric view. We also remove all movement actions from the action space, leaving only view-adjustment actions. As a result, the agent need not devise a search strategy or physically approach the target; it merely adjusts its gaze, enabling a focused evaluation of the model’s fine-grained spatial alignment capability.

\noindent \textbf{Navigation.} We define the navigation task as follows: Given a target object, the agent is deemed successful upon reaching within 1 meter of that object. To ensure sufficient path complexity, the agent is initialized at the corner of the room farthest from the target object. Meanwhile, the target object is guaranteed to remain visible within the agent's initial field of view, so that the challenge lies purely in the fundametal navigation capabilities.

\noindent \textbf{Planning.} The goal of this task is to evaluate an agent’s task-planning ability. In essence, this ability corresponds to the brain’s cognitive reasoning functions rather than the cerebellum’s motor-control functions. To effectively decouple motor control from planning, we abstract the four basic motion primitives into directly callable navigation interfaces. We adopt an interactive-task framework because the explicit, multi-stage nature of its execution process is especially well-suited for fine-grained evaluation of planning capability.

\subsection{Data Collection}
The data samples are gathered through a rigorous pipeline that comprises automatic sample generation and human-machine collaborative filtering, ensuring each sample is both high-quality and appropriately challenging.

By exploiting AI2-THOR’s comprehensive environment \cite{kolve2022ai2thor} and object metadata—such as 3-D coordinates, state flags, and instance-segmentation masks—we batch-generate candidate samples for every task. For the exploration task in particular, we first query the simulator to retrieve all objects in the scene along with their associated receptacles, and then automatically instantiate the four previously defined question types. 

To further enhance sample quality, we implemented a human-machine collaborative approach. We deployed an advanced MLLM to conduct 5 rounds of rollout evaluations on the samples, tracking the success rate for each sample. We then identified samples that either achieved complete success or complete failure across all rounds. Human experts subsequently intervened to assess the task feasibility of all-fail samples and the difficulty level of all-pass samples. We filtered out infeasible erroneous samples from the all-fail samples and excluded samples with insufficient difficulty from the all-pass samples, ultimately obtaining all samples for NativeEmbodied. More details of the data collection pipeline are provided in Appendix.

\subsection{Dataset Statistics}
Figure \ref{fig:dataset} illustrates our NativeEmbodied benchmark, which consists of 1,085 samples and encompasses three high-level tasks as well as four low-level tasks, making it a comprehensive evaluation tool.

\begin{figure}[t]
  \centering 
  \includegraphics[width=0.8\linewidth,keepaspectratio]{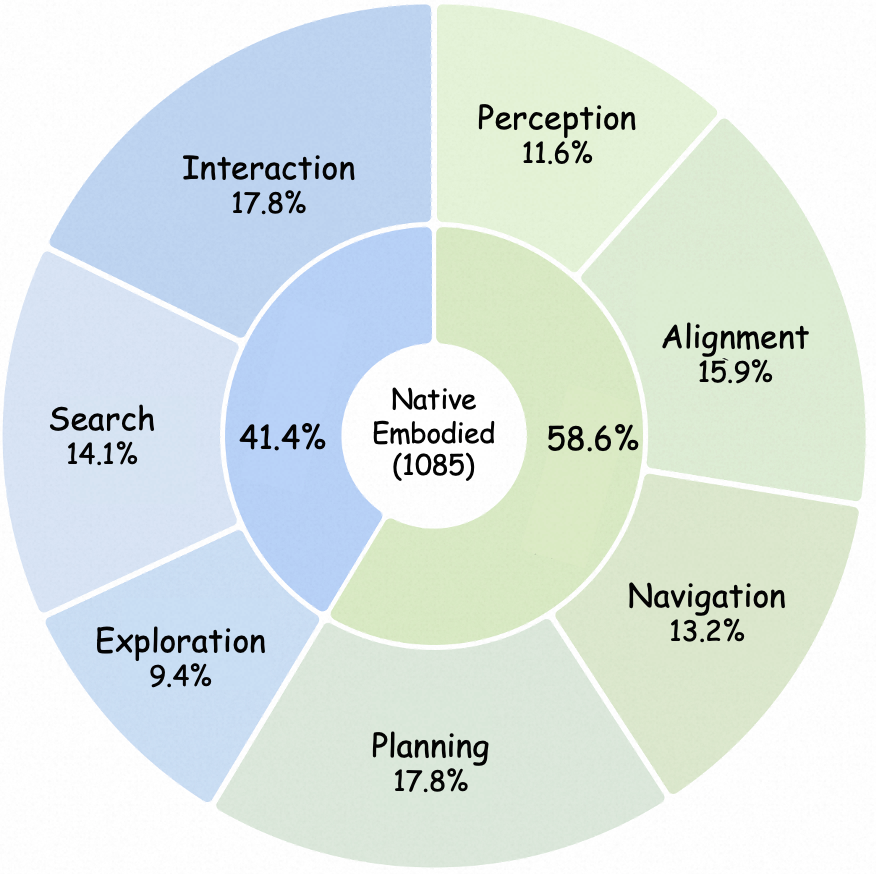} 
  \caption{Sample distribution of NativeEmbodied.}
  \label{fig:dataset} 
\end{figure}

\section{Experiment}

\subsection{Main Results}\label{sec:main}
\begin{table*}[ht]
\centering
\small
\label{tab:model-performance-revised}
\begin{tabular}{lccc|cccc|ccc}
\toprule
\multirow{2}{*}{\textbf{Model}} 
        & \multicolumn{3}{c|}{\textbf{Exploration}} 
        & \multicolumn{4}{c|}{\textbf{Search}} 
        & \multicolumn{3}{c}{\textbf{Interaction}} \\
\cmidrule(r){2-4}\cmidrule(r){5-8}\cmidrule(l){9-11}
        & Acc$\uparrow$ & AS$\downarrow$ & WAS$\downarrow$
        & SR$\uparrow$ & ACPD$\downarrow$ & AS$\downarrow$ & WAS$\downarrow$
        & SR$\uparrow$ & AS$\downarrow$ & WAS$\downarrow$ \\
\midrule
\multicolumn{11}{c}{\textbf{\textit{Closed-Source Large Vision Language Models}}}\\
\midrule
GPT-4o                    & 36.89 & 12.32 & 24.11 & 0.65 & 131.29 & 25.00 & 30.96 & 22.28 & 12.25 &26.84 \\
GPT-4v                    & 36.89 & 10.42 & 23.41 & 3.27 & 112.53 & 12.60 & 30.35 & 37.31 & \textbf{12.07} & \textbf{24.03} \\
GPT-o3                    & \textbf{52.43} & 11.06 & 20.54 & \textbf{34.64} & \textbf{32.94} & 15.60 & \textbf{25.67} & \textbf{38.34} & 13.35 & 24.25 \\
GPT-o4-mini               & 40.78 & 5.48 & 20.59 & 17.64 & 37.93 & 13.07 & 27.84 & 26.42 & 13.33 & 28.27 \\
Claude-3.5-sonnet         & 31.07 & 9.78 & 24.41 & 3.27 & 103.27 & 14.60 & 30.46 & 19.69 & 13.19 & 27.58 \\
Claude-3.7-sonnet         & 37.86 & 14.67 & 24.81 & 11.76 & 68.13 & 14.33 & 29.04 & 28.50 & 12.93 & 26.17 \\
Claude-4-sonnet           & 37.86 & 12.59 & 24.03 & 0 & 95.88 & - & 31.00 & 30.01 & 13.59 & 27.44 \\
Claude-4-opus             & 37.86 & 12.72 & 24.08 & 4.58 & 84.17 & \textbf{6.86} & 29.82 & 36.27 & 12.48 & 24.87 \\
Gemini-2.5-pro            & 40.78 & \textbf{4.71} & \textbf{20.28} & 14.38 & 35.89 & 7.91 & 27.68 & 33.68 & 12.17 & 24.67 \\
Gemini-2.5-flash          & 40.78 & 6.40 & 20.97 & 12.42 & 58.49 & 11.58 & 28.59 & 32.64 & 14.46 & 25.98 \\
Gemini-2.0-flash          & 39.81 & 11.51 & 23.24 & 2.61 & 90.83 & 14.75 & 30.58 & 24.87 & 13.53 & 26.79 \\
\midrule
\multicolumn{11}{c}{\textbf{\textit{Open-Source Large Vision Language Models}}}\\
\midrule
Qwen2.5-VL-72B            & 33.01 & 11.82 & 24.67 & 1.96 & 130.40 & 7.00 & 30.69 & 8.29 & 13.63 & 28.37 \\
Qwen2.5-VL-32B            & 31.07 & 14.6 & 25.41 & 1.31 & 129.93 & 23.00 & 30.83 & 6.74 & 13.15 & 29.61 \\
Qwen2.5-VL-7B             & 28.16 & 11.6 & 26.14 & 0 & 131.26 & - & 31.00 & 1.55 & 25.00 & 30.83 \\
Qwen2.5-VL-3B             & 25.24 & 8.13 & 26.03 & 0 & 131.68 & - & 31.00 & 0 & - & 31.00\\
\bottomrule
\end{tabular}
\caption{Performance comparison of closed-source and open-source LVLMs on the three high-level tasks: Exploration, Search and Interaction.  
For metrics, $\uparrow$ / $\downarrow$ mean "higher is better" / "lower is better".}
\end{table*}

\begin{table*}[t]
\centering
\small
\label{tab:model-performance-4tasks-spl}
\resizebox{\textwidth}{!}{
\begin{tabular}{lccc|cccc|cccc|ccc}
\toprule
\multirow{2}{*}{\textbf{Model}} &
\multicolumn{3}{c|}{\textbf{Perception}} &
\multicolumn{4}{c|}{\textbf{Spatial Alignment}} &
\multicolumn{4}{c|}{\textbf{Navigation}} &
\multicolumn{3}{c}{\textbf{Planning}} \\ 
\cmidrule(r){2-4} \cmidrule(r){5-8} \cmidrule(r){9-12} \cmidrule(l){13-15}
& P$\uparrow$ & R$\uparrow$ & F1$\uparrow$ 
& SR$\uparrow$ & ACPD$\downarrow$ & AS$\downarrow$ & WAS$\downarrow$
& SR$\uparrow$ & ACD$\downarrow$ & AS$\uparrow$ & WAS$\downarrow$
& SR$\uparrow$ & AS$\downarrow$ & WAS$\downarrow$\\
\midrule
\multicolumn{15}{c}{\textbf{\textit{Closed-Source Large Vision Language Models}}}\\
\midrule
GPT-4o            &  75.14& 73.15 & 74.28 &  7.51 & 86.85 & 3.91 & 15.07 & 50.00 & 2.16 & 6.87 & 11.42 & 58.55 & 9.63 & 14.82 \\
GPT-4v            & 79.51 & 78.11 & 78.83 &  6.94 & 66.81 & 3.23 & 15.12 & 55.56 & 2.23 & 7.81 & 11.43 & 62.18 & 9.25 & 14.04 \\
GPT-o3            & \textbf{83.15} & \textbf{84.51} & \textbf{83.97} & \textbf{64.16} & \textbf{22.73} & 7.28 & \textbf{10.4} & \textbf{63.19} & 2.02 & 8.34 & 11.08 & \textbf{72.54} & 10.71 & \textbf{13.87} \\
GPT-o4-mini       & 74.67 & 75.16 & 74.92 & 45.09 & 27.45 & 6.34 & 11.57 & 35.42 & 2.68 & 8.11 & 13.24 & 66.32 & 10.23 & 14.36 \\
Claude-3.5-sonnet & 76.59 & 72.33 & 73.82 &  9.83 & 63.38 & \textbf{2.82} & 14.72 & 47.92 & 2.01 & 7.83 & 12.12 & 55.44 & 10.40 & 15.55 \\
Claude-3.7-sonnet & 76.76 &73.27 & 74.35 & 20.23 & 60.91 & 4.01 & 13.62 & 42.36 & 2.14 & 7.92 & 12.55 & 60.62 & 11.47 & 15.84 \\
Claude-4-sonnet   & 77.51 & 73.58 & 74.77 & 36.41  & 29.39  & 6.63  & 11.83 & 27.78  & 2.43  & 4.39& 12.76  & 67.36 & 10.33 & 14.30 \\
Claude-4-opus     &81.21 & 81.14 & 79.59 & 39.31  & 28.74  & 7.87  & 11.28 & 53.47   & \textbf{1.72}  & \textbf{4.11} & \textbf{9.74}  & 67.88 & 10.35 & 14.16 \\
Gemini-2.5-pro    &  80.15 & 80.87 & 80.53 & 45.09 & 26.01 & 4.49 & 10.72 & 41.67 & 2.40 & 7.26 & 12.41 & 68.39 & 9.50 & 13.67 \\
Gemini-2.5-flash  & 77.98 & 79.47 & 78.42 & 35.84 & 30.36 & 7.41 & 12.93 & 38.19 & 2.65 & 7.67 & 12.78 & 52.33 & 10.54 & 16.35 \\
Gemini-2.0-flash  & 72.71 & 74.33 & 73.39 &  9.25 & 84.46 & 3.93 & 14.91 & 37.50 & 2.81 & 8.21 & 13.32 & 48.19 & 10.41 & 16.91 \\
\midrule
\multicolumn{15}{c}{\textbf{\textit{Open-Source Large Vision Language Models}}}\\
\midrule
Qwen2.5-VL-72B & 77.86 & 74.34 & 76.42 & 12.72 & 80.58 & 4.93 & 14.61 & 61.11 & 2.21 & 6.19 & 10.03 & 37.82 & 9.78 & 17.16 \\
Qwen2.5-VL-32B & 73.51 & 72.15 & 72.86 & 7.51   & 85.32  & 4.14  & 14.93   & 36.11 & 2.36  & 7.28 & 12.32  & 25.39 & 9.47 & 18.32 \\
Qwen2.5-VL-7B  & 71.61 & 70.74 & 71.01 & 5.78   & 86.14  & 3.33  & 15.12   & 25.00 & 2.71  & 7.21 & 12.81  & 12.95 & 10.28 & 19.56 \\
Qwen2.5-VL-3B  & 68.61 & 66.59 & 67.12 & 4.05   & 88.93 & 3.01  & 15.21   &  19.44 & 2.95  & 8.34 & 13.82  & 3.63 & \textbf{7.38} & 20.83 \\
\bottomrule
\end{tabular}
}
\caption{Performance of selected LVLMs on four low-level tasks: Perception, Spatial Alignment, Navigation and Planning.  
$\uparrow$ / $\downarrow$ denote “higher is better” / “lower is better”.}
\vspace{-0.3cm}
\end{table*}

\subsection{Evaluation Setup}
\paragraph{Baselines.} We evaluate 15 open-source and closed-source models, covering four model families: 
\begin{itemize}[leftmargin=*]
\item GPT family\footnote{\url{https://openai.com/index/}}: GPT-4o, GPT-4v, GPT-o3, GPT-o4-mini.
\item Claude family\footnote{\url{https://www.anthropic.com/news/claude-3-5-sonnet}}: Claude-3.5-Sonnet, Claude-3.7-Sonnet, Claude-4-Sonnet, Claude-4-Opus.
\item Gemini family \cite{team2024gemini}: Gemini-2.0-flash, Gemini-2.5-flash, Gemini-2.5-pro.
\item Qwen family\footnote{\url{https://help.aliyun.com/zh/model-studio/developer-reference/use-qwen-by-calling-api}}: Qwen-2.5-VL-72B, Qwen-2.5-VL-32B, Qwen-2.5-VL-7B, Qwen-2.5-VL-3B.
\end{itemize}

\paragraph{Environment.} In each turn of interaction, the agent receives a 640 × 480 egocentric image with a 90° field of view as input and outputs one action with specified parameters from the action space. The rollout step limits are 15 for alignment and navigation, 20 for planning, and 30 for three high-level tasks. The text–image history is truncated to 20 turns. During inference, the temperature of all VLMs is uniformly set as 0 to ensure reproducibility and consistency.

\paragraph{Evaluation Metrics.}
To obtain a more comprehensive and fine-grained picture of an agent’s performance, we report the following metrics in addition to \textbf{Success Rate (SR)}:

\begin{itemize}[leftmargin=*]
  \item \textbf{Average Steps (AS):}
        The mean number of steps taken in successful episodes,
        reflecting how efficiently the agent completes a task.

\item \textbf{Weighted Average Steps (WAS):}
    For each successful trajectory we use its \emph{actual} length,
    whereas for each failed trajectory we assign a penalised length equal
    to the task's predefined maximum number of steps $T$ plus a
    penalty factor~$\alpha > 0$ (set to 1 in our experiment).
    Formally, let
    $\mathcal{S}$ and $\mathcal{F}$ be the sets of successful and failed
    episodes,
    $s_i$ the number of steps taken in the $i$-th successful episode.
    The WAS is,
    \begin{equation}
      \mathrm{WAS}
      \;=\;
      \frac{
          \displaystyle\sum_{i \in \mathcal{S}} s_i
          \;+\;
          \displaystyle\sum_{j \in \mathcal{F}} (\alpha + T)
        }{
          |\mathcal{S}| + |\mathcal{F}|
        }.
    \end{equation}
    \noindent
    A smaller WAS indicates that the agent not only succeeds frequently
    but also does so efficiently.

  \item \textbf{Average Closest Distance (ACD):}
        The shortest Euclidean distance between the agent and the
        target object across the trajectories.

  \item \textbf{Average Closest Pixel Distance (ACPD):}
        The mean of the minimum pixel distance between the target
        object and the view center across the trajectories.
\end{itemize}

\noindent
We report
\textbf{Precision}, \textbf{Recall}, and \textbf{F1} score for \emph{Perception}. Each predicted triplet is considered positive only when all of its parts exactly match the ground truth.

\subsection{Main Results}\label{sec:main}
\textbf{High-level tasks in native settings pose significant challenges for VLMs.} Table 2 shows the performance of various VLMs on the three categories of high-level tasks. Even the strongest VLMs generally struggle with high-level tasks under native settings. This is particularly evident in Search, where the best-performing model GPT-o3 achieves only a 34.9\% success rate, while Claude-4-Sonnet fails to complete even a single task successfully. The same pattern holds for Interaction and Exploration, with the highest success rates being merely 52.4\% and 38.3\% respectively. This indicates that in native embodied environments, current VLMs are still far from being capable of effectively executing complex tasks.

\noindent \textbf{VLMs exhibit significant performance disparities across different low-level tasks.} As shown in Table \ref{tab:model-performance-4tasks-spl}, delving into various low-level tasks, we find that models demonstrate clear differentiation in performance across different tasks. First, VLMs display satisfying performance on Perception, indicating that current models' reliable capability in understanding and recognizing visual information. Second, in Planning, VLMs similarly demonstrate strong capabilities, with proprietary models generally achieving success rates exceeding 50\%, showcasing their reliability in task planning and reasoning. However, when tasks involve fine-grained operations in embodied environments, model performance shows a significant decline. In Navigation, more than half of the models achieve success rates below 50\%, with the worst-performing proprietary model achieving only 27.8\% success rate. Even more surprisingly are the results for Alignment—these seemingly simple operations have become a significant challenge for VLMs. Except for GPT-4o, no model achieves a success rate exceeding 50\%, and among closed-source models, four models achieve only single-digit success rates. These results clearly indicate that while VLMs have made progress in certain aspects, fundamental skill deficiencies still exist for specific basic embodied skills, particularly in tasks requiring dynamic spatial interaction.

\noindent \textbf{Mainstream VLMs exhibit distinct behavioral spectra in native setting.} The contrast between high- and low-level tasks not only reveals the limitations of each model but also allows them to demonstrate distinct strategic tendencies in embodied environments. In Navigation, GPT-o3 leads with the highest success rate, yet at the cost of significantly longer average step counts, revealing a robust and conservative path-planning preference. In contrast, Claude-4-Opus maintains over 50\% success rate with less than half the steps of GPT-o3, and leads in both ACD and WAS metrics, reflecting a more aggressive, efficiency-first exploration style. In Exploration, GPT-o4-mini and Gemini-2.5-Pro have significantly fewer average steps than other models, yet still achieve high accuracy rates second only to GPT-o3, indicating that both are more agile and confident in collecting and utilizing environmental information.

\begin{figure}[t]
  \centering 
  \includegraphics[width=\linewidth,keepaspectratio]{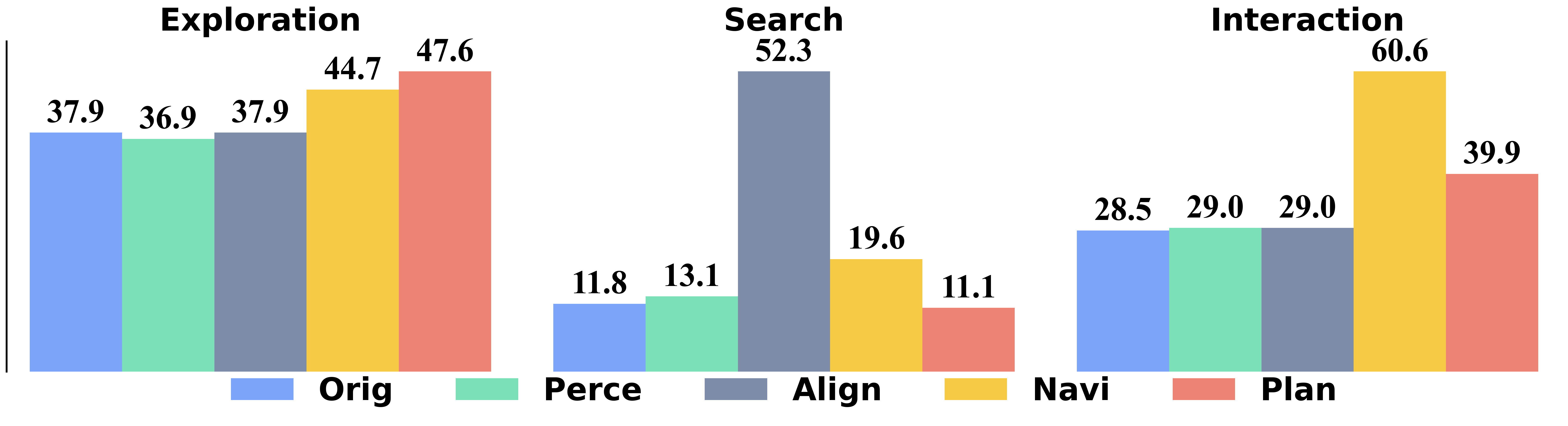} 
  \caption{Results from the skill-oriented ablation study, aimed to precisely identify the key atomic skills that limit model performance.}
  \label{fig:skill_ablation} 
\end{figure}

\subsection{Ablation Study of Foundamental Skills}
The main experimental results in Section \ref{sec:main} demonstrate that current VLMs exhibit significant limitations when executing complex tasks in native embodied environments. Basic skill assessments further reveal deficiencies in models' core embodied capabilities. To precisely identify the key atomic skills limiting model performance, we conducted systematic skill ablation experiments:
\begin{itemize}[leftmargin=*]
  \item Perception: We utilize AI2THOR's API to extract instance segmentation from each egocentric image, converting it into structured text descriptions through predefined templates as supplementary input to the model.
  \item Alignment: A ``LookAt'' is provied for the agent to directy aim view into the target object if visible.
  \item Navigation: The agent is allowed to teleport to the target object if visible.
  \item Planning: We pre-decompose tasks into subtask sequences, asking the model to execute them step-by-step.
\end{itemize}
We selected Claude-3.5-Sonnet as our experimental subject, as this model demonstrates moderate performance in benchmark tests, offering good representativeness that facilitates more generalizable conclusions. As shown in Figure~\ref{fig:skill_ablation}, the experimental results reveal three important insights:

\noindent \textbf{Mature Perception Capabilities.} The introduction of ground-truth perception information failed to significantly improve model performance,indicating current advanced VLMs already possess sufficient visual capabilities.

\noindent\textbf{Dual Bottlenecks in Long-Horizon Tasks} In Exploration and Interaction tasks, ablation on both planning and navigation yield significant improvements, indicating that both cognitive-level decision-making abilities (planning) and action-level execution abilities (navigation) are key bottlenecks for long-horizon tasks. 

\noindent \textbf{Fine-Grained Spatial Requirements in Search Tasks} In Search tasks, improvements to navigation and alignment capabilities (particularly alignment) showed significant effects, while planning capability had limited impact. This reflects the unique characteristics of search tasks: their immediate-response nature reduces dependence on complex planning, but demands extremely high precision in spatial positioning and viewpoint control.

\begin{figure}[t]           
  \centering                  
  \includegraphics[width=\linewidth,keepaspectratio]{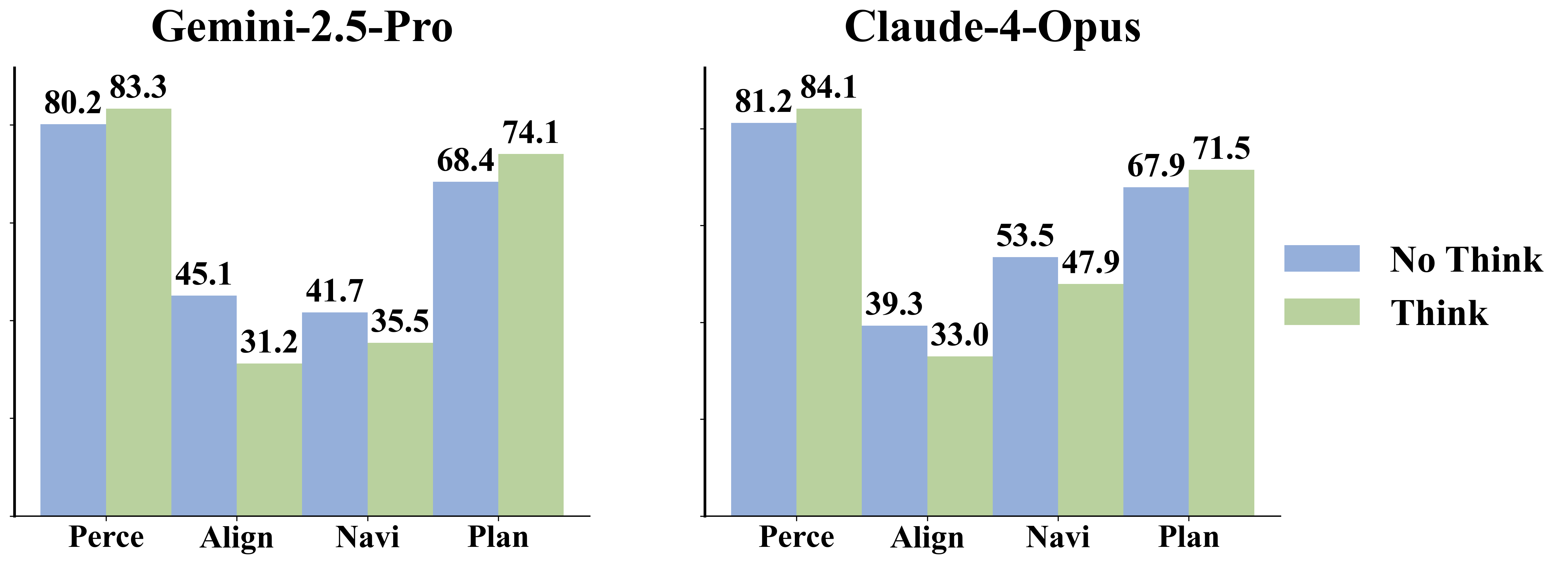} 
  \caption{Results of think mode ablation study, aimed to explore the capabilities of reasoning models.}
  \label{fig:think_ablation}          
\end{figure}
\subsection{Ablation Study of Think Mode}
Reasoning models' improved problem-solving skills \cite{huang2025visionr1,liu2025visualrft,gu2025mobile} motivate us to explore whether it could help break through current bottlenecks. We selected two specialized reasoning models: Gemini-2.5-Pro and Claude-4-Opus. Specifically, in each round of rollout, we first ask the models to think\footnote{Notably, for reasoning models, we enable their reasoning mode; for non-reasoning models, we request them to output their thinking process in the prompt}, then select an action. The experimental results are shown in Figure~\ref{fig:think_ablation}, from which we can draw the following insights:

\noindent \textbf{Thinking enhances cognitive abilities for embodied environments}. After enabling thinking mode, the overall preformance on both perception and planning tasks increased, indicating that the reasoning process helps models better understand environmental states, identify key elements, and formulate more reasonable high-level strategies. This improvement is particularly pronounced in tasks requiring complex reasoning and long-term planning.

\noindent \textbf{Thinking may interfere with basic action execution}. After engaging in thinking mode, success rates for tasks that require precise action actually decreased significantly, such as alignment and navigation. This decline might be attributed to excessive reasoning processes, which can introduce unnecessary complexity and interfere with the intuitive execution of basic actions.

These findings reveal the double-edged sword effect of  introducing reasoning capabilities into embodied agents: while reasoning can enhance cognitive abilities, it may interfere with low-level motor control. This suggests that when constructing embodied agents, we need to more carefully balance the functional division between the ``cerebrum'' (cognitive reasoning) and ``cerebellum'' (action control).

\subsection{Error Case Analysis}
\begin{figure}[t]           
  \centering                  
  \includegraphics[width=0.9\linewidth,keepaspectratio]{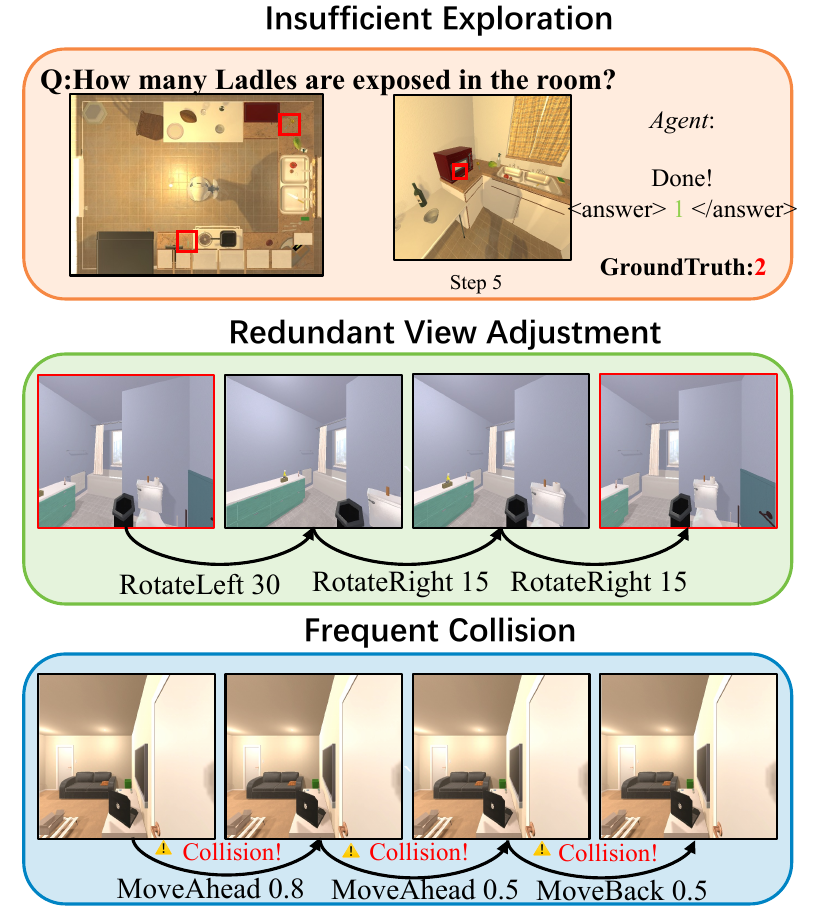} 
  \caption{Case study of common error trajectories .}
  \label{fig:error}          
\end{figure}

As shown in Figure \ref{fig:error}, we summarized three categories of the most common errors exhibited by agents evaluated in our NativeEmbodied benchmark:
\begin{itemize}[leftmargin=*]
\item \textbf{Insufficient Exploration}: Agents sometimes fail to conduct comprehensive exploration of the environment, prematurely drawing conclusions based solely on partial information, demonstrating overconfidence.
\item \textbf{Redundant View Adjustment}: Agents frequently perform repetitive and unnecessary view adjustments within a considerable number of valid steps, severely reducing task execution efficiency. Worse still, the resulting repetitive observations can sometimes lead agents into operational dead loops. 
\item \textbf{Frequent collision}: Agents exhibit poor perception and response to environmental collisions, unable to make effective adjustments based on historical information, resulting in frequent collisions. This issue is particularly severe when agents are in confined spaces such as corners, where they easily become stuck and unable to escape.
\end{itemize}

\section{Conclusion}
In this work, we presented NativeEmbodied benchmark, a comprehensive benchmark for evaluating VLM-driven embodied agents using a unified, native low-level action space. Through systematic evaluation of both low-level and high-level tasks across 15 open-source and closed-source VLMs, we identified significant limitations in fundamental embodied capabilities that directly impact performance on complex tasks. Our findings not only highlight the current challenges in VLM-driven embodied intelligence but also provide valuable guidance for future development in this field.

\clearpage
\newpage
\section{Acknowledgements}
This work was supported by the grants from National Natural Science Foundation of China (No.62222213, 62072423).
\bibliography{aaai2026}

\end{document}